\documentclass[11pt]{article}

\usepackage{graphics}
\usepackage{graphicx}
\usepackage{amsmath}
\usepackage{bbm}
\usepackage{amssymb}
\usepackage{amsmath}
\usepackage{url}
\usepackage{color}
\usepackage[centering, margin={1in, 1in}, includeheadfoot]{geometry}
\usepackage{epstopdf}
\usepackage{dsfont}
\newcommand{\E}{\mathcal{E}}
\newcommand{\G}{\mathcal{G}}
\newcommand{\D}{\mathcal{D}}

\newcommand{\I}{\mathcal{I}}
\DeclareMathOperator*{\argmin}{arg\,min}

\newcommand{\mS}{\mathcal{S}}
\newcommand{\N}{\mathcal{N}}

\usepackage{titlesec}
\titleformat*{\section}{\large\bfseries}
\titleformat*{\subsection}{\normalsize\bfseries}
\titleformat*{\subsubsection}{\small\bfseries}

\begin{document}
%\linenumbers
%\setlength\linenumbersep{10pt}
%\renewcommand\linenumberfont{\normalfont\tiny\sffamily\color{blue}}
% paper title
% can use linebreaks \\ within to get better formatting as desired
\title{Robust Topological Feature Extraction for  Mapping of Environments using Bio-Inspired Sensor Networks}
\author{
{Alireza Dirafzoon and Edgar Lobaton  \footnote{ \noindent Department of Electrical Engineering, {North~Carolina~State~University}. Email: 
 \{adirafz, alper.bozkurt, ejlobato\}@ncsu.edu} }}
\date{}
\maketitle

\begin{abstract}
%\boldmath
 In this paper, we exploit  minimal sensing information gathered from  biologically inspired sensor networks to perform exploration and mapping in an unknown environment. A  probabilistic  motion model  of mobile sensing nodes, inspired by  motion characteristics  of cockroaches,  is utilized to extract  weak encounter information in order to build a  topological representation of the environment. 
  Neighbor to neighbor interactions among the nodes are exploited to build point clouds representing  spatial features of the manifold characterizing the environment based on the sampled data. 
 To extract dominant features from sampled data, topological data analysis is used to produce  persistence intervals for features, to be used for topological mapping.  In order to improve robustness characteristics of the sampled data with respect to outliers, density based subsampling algorithms are employed.  Moreover, a  robust scale-invariant  classification algorithm for  persistence diagrams is proposed to provide a quantitative representation of desired features in the data. Furthermore, various strategies for defining  encounter metrics with different degrees of information regarding agents' motion are suggested to enhance the precision of the estimation and classification performance of the  topological method. 
\end{abstract}
% no keywords
\newpage
\section{Introduction}
 %~~~~~~~~~~~~~~~~~~~~~~~~~~~~~~~~~~~~~~~
   \begin{figure}[b]
         \centering
       \includegraphics[width=0.9\linewidth]{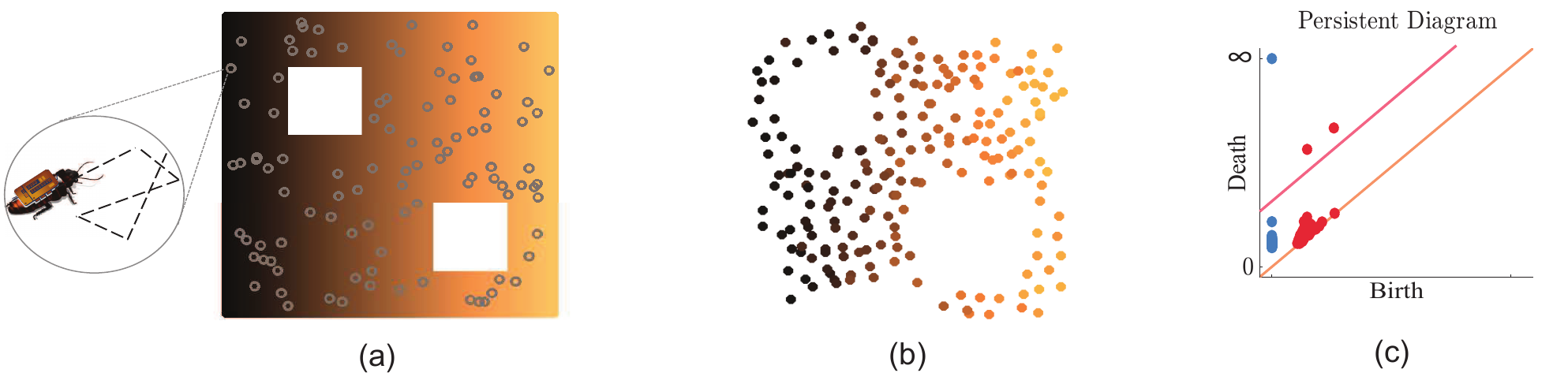}
       \caption{\small{Topologcal Mapping: (a) a physical environment with mobile sensing nodes moving inspired by cockroaches, (b) an esimated point cloud from coordinate free information,  (c) the corresponding persistence diagram highlighting the features in dim 0 (blue dots) the features in dim 1 (red dots); significant features can be distinguished from noise using an appropriate threshhold. }}
       \label{env}
    \end{figure}
     %~~~~~~~~~~~~~~~~~~~~~~~~~~~~~~~~~~~~~~~
Sensor networks with their broad application in  %surveillance,
  mapping and navigation  \cite{Kotay06}, habitat monitoring  \cite{Mainwaring2002}, exploration, and search and rescue  \cite{Kantor2003},  have attracted a lot of attention in recent decades.   Mobile sensor networks 
%  in which the sensing nodes are equipped with mobility,
    offer the flexibility to adapt with dynamic environments. As an example,  swarm robotic systems,  where  mobility models of agents are inspired by biological entities, and each agent is equipped with some type of sensing, are considered as  mobile sensor networks which can perform distributed sensing and estimation tasks. Emergent behavior in animals such as formation\cite{balch99}, coverage\cite{cortes04}, and aggregation\cite{gazi03}  have inspired scientists  to develop behavioral-based distributed systems.  
  
 In some applications, however,  the amount of information that could be sensed or transferred by the agents is limited.  This motivates the design of distributed systems composed of simple agents with  minimal  sensing requirements\cite{gandhi08}. 
  Consider for example a disaster zone response scenario, where we aim to perform exploration and  mapping of  an unstructured and unknown environment using a mobile sensor network. One may choose to make use of a swarm of biologically inspired agents with minimal sensing capabilities to perform the task.  However, under such rough conditions of the terrain, localization information provided by the agents could be very weak and contains a high amount of uncertainty. Hence, traditional localization and mapping algorithms such as SLAM\cite{SLAM} would fail to perform effectively. 

  Methods from computational topology, on the other hand, can provide tools to 
  %can  provide topological mappings while  not requiring  localization of agents, 
  extract topological features from data sets without requiring coordinate information. This makes them more suitable for scenarios in which weak or no localization is provided.
    % equipped with  complex locomotion capabilities (such as climbing and crawling \cite{XRL}), that enables them to explore and survive in such harsh environments. \\
% low cost sensors  for distributed sensing and estimation .... can be followed by more expensive  ... 
% some scenarios, it is impractical to ....
%\cite{Akyildiz2002} wireless sensor networks a survey\\
%sensor nodes consist of sensing, data processing, and communicating components   + random
%deployment in inaccessible terrains or disaster
%relief operations. + processing abilities to  carry out computations \cite{Akyildiz2002} \\
%
%%~~~~~~~~~~~~~~~~~~~~~~~~~~~~~~~~~~~~~~~~
%\begin{figure}[tbp]
%\begin{center}
%%\includegraphics[width=0.5\linewidth]{bar-dist.eps}  \\
%\end{center}
%\caption{Mobile Sensor Network}  \label{int}
%\end{figure}
%%~~~~~~~~~~~~~~~~~~~~~~~~~~~~~~~~~~~~~~~~~
Topological data analysis (TDA), introduced in \cite{Edel2000},  has been a new field of study which employs tools from persistent homology theory \cite{Edels08}
to obtain a qualitative description of the topological attributes and visualization of  data sets sampled  from  high dimensional point clouds. The point cloud can be thought of as finite samples taken from a density map which may include noise.  
%[computational geometry and topology as new techniques for data analysis].
%  ...  
TDA
%which could be thought of as a new scheme in  clustering,  exploits algorithms in computational topology to 
represents  the prominence of features in the point cloud in terms of 
a compact representation of the multi-scale topological structure
 called persistence diagrams\cite{Edels10}. It reduces the dimensions of data by construction of a filtration of  combinatorial objects, which can represent geometrical and topological features of the data set at specific scales.
%  , and the nodes at this complex can be considered as corresponding clusters in the data. 

 %.... using persistent homology to extract  topological features in data sets.
 %It will provide the number of locally or globally dominant  features of a density map a point cloud data,  ... 
% TDA recently has been used in a variety of applications including image processing and computer vision [carlson],  gene expression data, text mining [], elevation for protein docking[].
% Furthermore, 
% topological methods have been proposed to study coverage characterization and hole detection for  static sensor networks when no localization information is provided. 
Topological frameworks have been used for  characterization of coverage and hole detection in stationary sensor networks by using only proximity information of the nodes  within a neighborhood\cite{Ghrist, Silva2007}. 
% E. Lobaton et al.  \cite{Lobaton2010}  extended these topological models to incorporate long-range directional sensors in the presence of occluding objects, and demonstrated their approach in real networks of cameras.
%  
However, these studies mostly focused on networks with static nodes. Moreover, they are  mainly concerned about the coverage holes in the sensing domain of the  network rather than the  topology of the physical environment itself. 
%Persistent homology has been used to extract coverage hole information out of stationary sensor networks in \cite{Ghrist}. 
Adding mobility to the nodes in the network makes the problem more challenging. One way to reduce such complexity is to look for patterns created by tracing the encounters of the nodes instead of investigating the mobility data itself. Walker \cite{Walker2008} employed persistent homology to compute topological invariants from encounter data of the mobile nodes in Mobile Ad-Hoc networks in order to infer global  information regarding the topology of a physical environment, but the nodes are assumed to follow a simple mobility model on a graph.  

\noindent\textbf{Contribution:}
In this paper, we aim to construct topological maps of unknown environments using bio-inspired mobile sensor networks under the constraint of limited sensing information. In particular, we  consider agents whose movement model is described by 	the motion model of cockroaches, which are  experts to survive in harsh environments.
%, which includes  diffusive random walk and wall-following  behaviors. 
We develop algorithms that do not depend  on any type of traditional localization schemes. Instead, we consider estimation of a topological  model for the environment based on limited information retrieved from the agents including their  own status, and their encounters with other agents in their proximity. We propose various types  of metrics for the construction of point clouds based on  coordinate-free information that  estimate topological features of the environment.
% [ToDo: use of  parameters from the motion models as external inputs for efficient mapping of the environment],
% 
Moreover, density based subsampling methods are used to deal  with outliers produced as a result of uncertainty in the estimated point clouds. Computational topology  algorithms are used to extract dominance  of these features in terms of persistence intervals. 
% * estimate the topological features of an unknown environment with minimal sensing.
% especially, homology of the environemnet,
However, such intervals  provide qualitative representation  about low dimensional structures  in the point cloud. Hence, methods from  machine learning and statistical data analysis recently have been integrated with persistence analysis in order  to make inference and interpretation  out of such intervals \cite{Munch2013,Bal13}.
%For example, , mean and variance of a persistence diagrams are used  to model variations and obtain statistical properties in data. 
In this study, we develop a density based classification algorithm to extract features from persistence intervals quantitatively in a robust manner to be used in the construction of the topological map. 

The remainder of the paper is organized as follows: A concise background on TDA is presented in section \ref{math}. Bio-inspired mobility model as well as sensing  model for the nodes are described in section \ref{prob}, followed by a general overview of the proposed mapping framework. Section \ref{enc-met} discusses the different metrics employed for construction of point clouds. Subsampling techniques are described in section \ref{subs}, and classification algorithm for dominant features is proposed in section \ref{class}. 
% Section \ref{relw} provides an overview of the related previous work.  In Section \ref{model}, we describe the motion model of agents based on the natural behavior of cockroaches. The mathematical tools for the analysis and estimation of the topological models are  introduced in Section \ref{math}. In  Section \ref{results}, 
%simulation studies are presented including the exploration of a swarm control mechanism based on switches on motion behavior. 
%exploration and analysis of the application of the existing mathematical tools towards an effective estimation of the environment as well as our switching swarm controller are presented,  and simulation results are provided to support the discussion. 
Finally, conclusion and future extensions of the presented work are discussed in Section \ref{conc}.

\section{Background on TDA} \label{math}

%\subsection{ Topological Data Analysis} 
A brief introduction of some of the basic concepts in computational topology %and persistent homology 
%used for data analysis in this paper 
is presented here.  A comprehensive review of the  topic can be found in \cite{Edels10}. 
%~~~~~~~~~~~~~~~~~~~~~~~~~~~~~~~~~~~~~~~~~
  \begin{figure}[tb]
      \centering
   \includegraphics[width=0.88\linewidth]{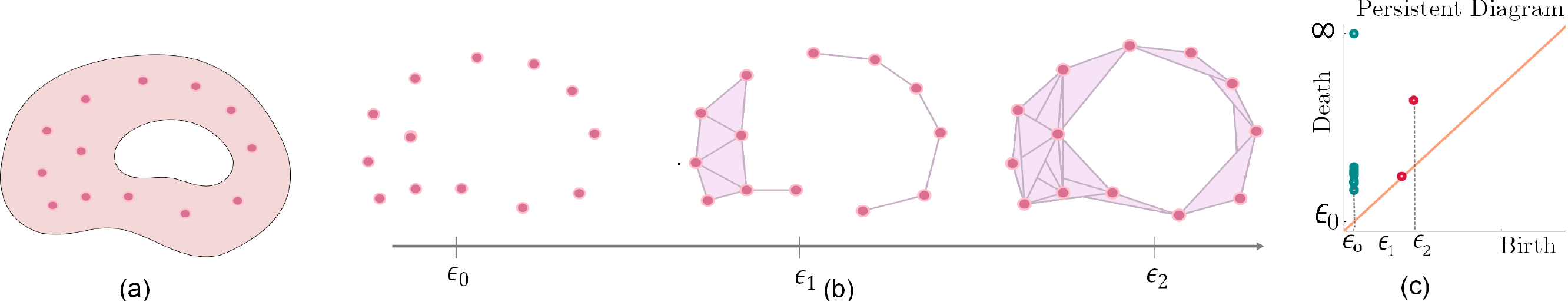}
%      \includegraphics[width=0.15\linewidth]{figs/Filt-pc} \includegraphics[width=0.2\linewidth]{figs/Filt1}    \includegraphics[width=0.2\linewidth]{figs/Filt2}
%    \includegraphics[width=0.18\linewidth]{figs/ComtopPers}
%     % \includegraphics[scale=1.0]{figurefile}
      \caption{ \small {Topological persistence: (a) An example of a topological space $ M $ with a sampled point cloud $ X $, (b) a filtration of complexes over $ X $, (c) the corresponding persistence diagram ( dgm$ _0(X) $ in dark cyan and  dgm$ _1(X) $ in red) }}
      \label{top}
   \end{figure}
 %~~~~~~~~~~~~~~~~~~~~~~~~~~~~~~~~~~~~~~~~~
One of the well-known techniques widely used in topological data analysis is persistent homology, 
%which in contrary to geometric methods that aim to estimate  shapes and distances, 
which deals with the way that objects are connected. Topological structures of a  space $M$ are summarized 
as a compact representation in the form of so-called  Betti numbers, which are ranks of  topological invariants, called \textit{homology groups}. The $n$-th Betti number, $ \beta_n $ measures the number of $n$-dimensional cycles in the space (e.g. $\beta_0$ is the number of connected components and $\beta_1$ is the number of  \textit{holes} in the complex).
The space $M$ usually is not directly accessible but  a  sampled version of it, ${X,}$ can be used for computations. This sample  is represented as a point cloud,  a finite set of  points equipped with a metric, which can be defined by pairwise distances between the points. 
% (distance function). The distance function can be defined based on the coordinates of the points in their embedding manifold, or can be constructed by specifying pairwise distances between points independently from their (possibly unknown) coordinates. 
% 

A standard method to analyze the topological structure of a point cloud  is to map it  into  combinatorial objects called {\it  simplicial complexes}. 
% in order  is to represent them as a {\it  simplicial complex}.
%Given a set of vertices $\mathcal{V}$, a $k$-simplex is defined as a set $\{v_1, v_2, \ldots, v_{k+1}\}$,   $v_i \in\mathcal{V}$,  $ \forall i $ and $ i \neq j,  \forall i,j$.   For example, a $0$-simplex is a vertex, a $1$-simplex is an edge, and a $2$-simplex is a triangle. 
%%A $k-1$-simplex $S_1$,  is called a face of a $k$-simplex $S_2$,  if the vertices of $S_1$ form a subset of the vertices of $S_2$ . 
%%A $k-1$-simplex $S_1$   is called a face of a simplex $S_2$,  if the vertices of $S_1$ area  subset of the vertices of $S_2$ . 
%A finite collection of simplices,  called a \textit{simplicial complex}  if for each simplex, each of its faces are also included in the collection. 
%if whenever a simplex lies in the collection then so does each of its faces. 
%Simplices are defined as  combinatorial objects whose vertices are just labels requiring no coordinates in space, and a collection of these objects constructs  a simplicial complex.  
%A simplicial complex representation of the objects in figure \ref{top} is shown in figure \ref{}.
One way to build these complexes is to select a scale $\epsilon$, place balls of radius $\epsilon$ on each vertex, and construct simplices based on their pairwise distance relative to $\epsilon$. A computationally efficient complex, called Vietoris-Rips complex \cite{Erin10} consists of simplices for which the distances between each pair of its vertices are at most $\epsilon$. 
A sequence of complexes, called a filtration ${X}(\epsilon)$,  then can be obtained by increasing $\epsilon$ over a range of interest,  with the property that 
if  $ t< s $ then  ${X}(t)\subset {X}(s)$. 
% meaning that all simplices in $X(t)$ are included in $X(s)$. 
Persistent homology, computes the  values  $\epsilon$ for which the classes of topological features  appear ($b^i_n$) and disappear ($d^i_n$) during filtration,   referred to as  the birth and death values of the $i$-th class of features in dimension $ n $. 
 This information is encoded into persistence intervals $[b^i_n, d^i_n] $ or as a multiset of points $ (b^i_n, d^i_n) $, called a persistence  diagram, dgm$_n $($ {X} $). Algorithms for computation of persistent homology  can be found in \cite{Edel2000, Zl04}. 

An example of a topological space $M $ together with sampled data ${X} $ and corresponding filtration over $ \epsilon $ is shown in figure \ref{top}. The persistent diagram infers the existence of one connected component and one persistent hole. 

\section{Problem Formulation}\label{prob}

Consider a network of  mobile sensing agents in a bounded environment $\mathcal{D} \in \mathbb{R}^d$, each distinguished with their unique  ID's. We assume that motion dynamics of the agents  in a bounded space mimic the movements of cockroaches,  as described in the next subsection. 
Moreover, we assume that  the agents are provided with weak localization information, i.e. they can only identify the other agents within a certain radius, and no coordination information is provided, which is described in the following subsection. 
%However, they are provided with their own tags, and are able to capture the tags of the other agents coming into their detection radius, as well as the time at which this encounter event occurs. This information is sent to a base station.  

%\begin{figure}[tpb]
%        \centering
%      \includegraphics[width=0.5\linewidth]{figs/env2}
%      \includegraphics[width=0.5\linewidth]{figs/swarm}
%      \caption{\small{..}}
%      \label{rw}
%   \end{figure}

\subsection{Bio-Inspired Mobility Model}\label{model}
The mobility  model is adopted from the probabilistic movement model of cockroaches described in \cite{Jeanson2005}.  Motion characteristics of the mobility model of the cockroaches can be mainly described by their individual and group behaviors. In this paper, for the sake of simplicity, we skip the group behavior resulting  from interactions among individuals. 

% [we will use terms agents and cockroaches alternatively afterwards ] 
The  individual behavior of  cockroaches depends on their relative distance to the boundaries of the arena.  
Specifically, when they are far from the boundaries of the environment, their motion  can be modeled as 
a diffusive \textit{random walk} (RW) with piecewise fixed orientation movements, characterized by line segments, interrupted by direction changes, and constant average velocity of $v_c$. The average length of the line segments $l^*$, is considered as the characteristic length of an exponential distribution for the path lengths. As for angular reorientation, we assume an isotropic diffusion, where  $p(\theta_{\text{new}} |\theta_{\text{current}} )   =  p(\theta_{\text{new}} )  $ characterized by a uniform distribution over $  (0, 2\pi)$. 
%The agents are assumed to move with a constant average velocity of $v_c$. 
%  resulting  in  a uniform distribution of angle reorientation independent of the previous angle. 
% % % % % %    WF % % % % % % % % % % % % % %
 %The movement model of an individual agent could be obtained by incorporating the motion models in the two regions.
  
%When the agents enter the $\mP$ partition, 
   %are close enough to the boundary of the environment, 
  When cockroaches detect a part of the boundary of  $ \D $ (through their antennas)  they  switch to a \textit{wall following}  (WF) behavior with a constant average velocity of $ v_p $. 
  During wall following, the agents might  leave the boundary towards the central partition randomly  after a random time modeled with an exponential distribution with an average of 
   $\tau_{\text{exit}}$ seconds, 
%    which denotes the  characteristic time of  the corresponding exponential distribution. 
  %with a  probability $p_{\text{exit}} = \delta t/ \tau_ {\text{exit}}$, where
%They leave the peripheral partition 
and an  angle of $\theta_{\text{exit}}$  with respect to the tangent vector to the boundary distributed uniformly over $[0, \pi]$.
 % As in  the  central  partition,  the agents stop for a short o long rest  on average after a characteristic time of 
   %  the  agents  stop during their random walk in average after a characteristic time of
%     $\tau_{\text{stop,p}}$.
   
   However, during their RW or WF movement, the agents  probabilistically \textit{stop} for some period of time and then continue their movement. The agents' stopping behavior  is  modeled as a memory-less process described  by an exponential distribution with a characteristic time $\tau_{\text{stop}}$,  representing the average time elapsed before an agent stops. 
   %with a probability of $p_{\text{stop,p}}  = \delta t/\tau _{\text{stop,p}}$ , where $\delta t$  denotes the time interval and    the  of and denotes . 
Once an agent enters a stop mode (S), it could either have a  short stop with a probability of $p_{\text{sh}}$, or a long stop with the probability of $ 1 - p_{\text{sh}}$. Each  of  these  two  stopping  processes  are characterized by exponential distributions with characteristic times $ \tau_{s}$  and $\tau_{l}$, respectively.
%\tc{blue}{Assumption that we are able to control their behavior to switch from a mixed random walk wall following to pure random walk  (RW) or wall following (WF) or a sopped status (S) (Force Stop). }
More details on the probabilistic motion model and the parameters we used can be found in \cite{Dirafzoon2013}.

  %~~~~~~~~~~~~~~~~~~~~~~~~~~~~~~~~~~~~~~~~~
%\begin{table}[!b]
%\caption{{Random walk and wall following  model parameters } }
%\label{partab1}
%\begin{center}
%\begin{tabular}{c|c|c|c|c|c|c}
%Zone & Mean Speed  & $l^*$   & $\tau_\text{stop}(n)$ & $p_\text{sh}(n)$ & $\tau_s(n)$ & $\tau_l(n)$ \\
% \hline 
% C       & 			15 mm/sec	         &  30 mm	 & 20&  0.93 & 0.5& 20  \\
%P  &  	  10 mm/sec           &   - & 20&  0.93 & 0.5& 20\\
%\end{tabular}
%\end{center}
%\end{table}
%~~~~~~~~~~~~~~~~~~~~~~~~~~~~~~~~~~~~~~~~~

% As in  the  peripheral  partition,  the agents stop their random walk   on average after a characteristic time of 
%  the  agents  stop during their random walk in average after a characteristic time of
 % $\tau_{\text{stop,c}}$.
%  a probability of $p_{\text{stop,c}}  = \delta t/\tau _{\text{stop,c}}$ with   being . 
  %The short and long stops are characterized with  probabilities of $p_{\text{sh,c}}$   and  $ 1 - p_{\text{sh,c}}$  and characteristic times $ \tau_{s,c}$  and $\tau_{l,c}$, respectively.

\subsection{Sensing and Communication }

Likewise the mobility model, the sensing model is inspired by limited sensing capabilities of cockroaches combined with capacities  added by integration of wireless transmitter and receiver chips inserted into their body \cite{Latif2012}.
 Specifically, the nodes can detect other agents as well as boundaries of the environment within a detection radius of $ r_d $. Each agent is equipped with  a unique ID and is able to record and transmit its own ID, the ID's of the other agents in its detection neighborhood, and corresponding time of the occurrence of such encounters to the base station. Furthermore,  the nodes are  able to report their status as being in a RW, WF, or S state. 
%, communication, and computational features of our sensor network model. \\
%\textcolor{red}{Computational power ... }\\
%\textcolor{red}{Communication Model: ??}
%
%
\subsection{Algorithm Overview}
\begin{figure}[tb]
      \centering
      \includegraphics[width=0.7\linewidth]{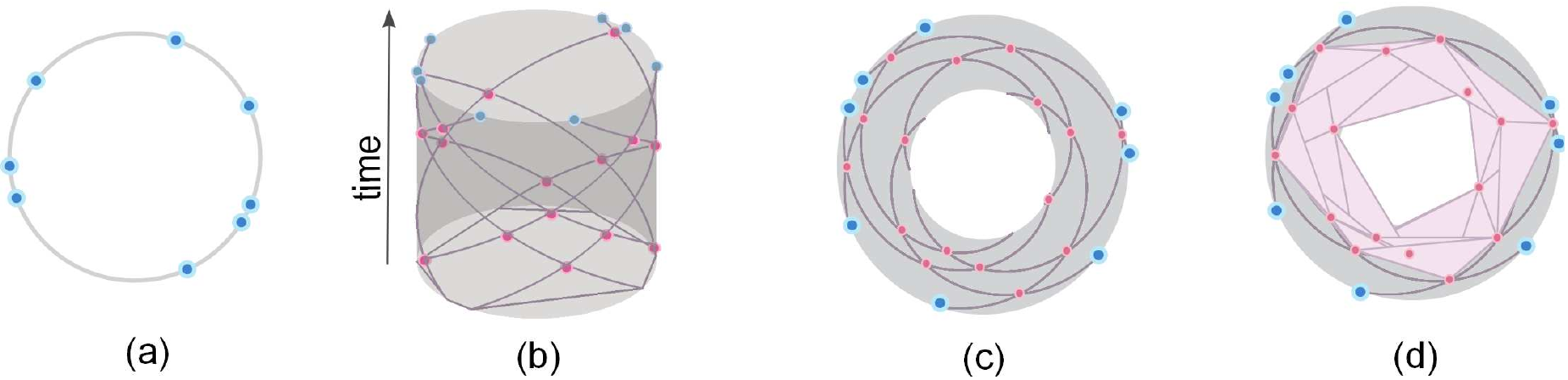}
      \caption{\small {Construction of encounter complex: (a) Nodes (blue dots) moving on a circle $ S^1 $ and (b) their encounters over time (pink dots) in the encounter space (cylinder $ S^1 \times \mathbb{R^+} $); (c) nodes moving on an annulus and (c) their corresponding encounter complex }}
      \label{enc}
   \end{figure}

%
% \begin{figure}[tb]
%      \centering
%      \includegraphics[width=0.85\linewidth]{figs/block}
%     % \includegraphics[scale=1.0]{figurefile}
%      \caption{\small { Block Diagram of Mapping Algorithm  }}
%      \label{block}
%   \end{figure}
%~~~~~~~~~~~~~~~~~~~~~~~~~~~~~~~~~~~~~~~~~
 %~~~~~~~~~~~~~~~~~~~~~~~~~~~~~~~~~~~~~~~~~
%
The overall  mapping algorithm is summarized as the following steps:\textit{ Exploration}, \textit{Data Collection}, \textit{Computing Encounter Metric}, \textit{Subsampling}, \textit{Topological Analysis and Visualization}, and \textit{Classification}. 
%in figure \ref{block}. 
Exploration of the unknown environment is performed by  the agents based on the
probabilistic motion model described in Subsection \ref{model}.
% Necessary  switching could be performed among NM, RW, WF, and FS behaviors according to the \tc{red}{? which will be discussed later .. }.
Minimal sensing in the agents with no odometry information along with bandwidth and power constraints in harsh environments, results in a lack of coordinate information for the purpose of environment mapping. 
  However, for a coordinate-free mobile network,  data associated with the encounters between agents can be used instead of dealing directly with 
coordinates of moving nodes.
 Each agent sends its local encounter information to the base station, where this information is processed to construct a metric, which we refer to as the  estimated \textit{encounter metric}, and the corresponding \textit{encounter point cloud}  is built on the set of  nodes.  
 The  point cloud is processed to construct a filtration of simplicial  complexes, denoted as \textit{encounter complexes}.  
 Figure \ref{enc} illustrates  construction of encounter complex for a simplified motion of 8 agents over circular regions. 

Extracting topological information from the whole point cloud would be computationally expensive due to the large number of events that are created over time; hence a subsampling algorithm needs to  be used  to reduce the computational complexity. 
A subset of the point cloud is selected such that it possesses  the same topological properties as the original set, and exploited to construct a smaller filtration of complexes. Persistent homology  is then used to extract ordinary persistence intervals for subsampled data. These intervals are then employed for feature extraction towards building the environment map.  We used Rips complexes for construction of filtrations, and Dionysus C++ library \cite{Dion2} for computation of persistent homology. 
Finally our proposed density based classification  technique is used  for  robust feature extraction from  the estimated point cloud. For visualization purposes, we used 
Multi-Dimensional Scaling (MDS)\cite{mds} to obtain  projections of the point cloud  on 3D Euclidean space. 

% for large datasets (in our case when there are  a lot of encounter events created by probabilistic movements of insect-like agents),  using the whole set of vertices  results in too large complexes, which are computationally very expensive. 
%%  This brings out the main idea behind the witness complex\cite{Silva2004},
%Hence,  usually a small subset of the points, called \textit{landmark points}, is selected on which  a smaller filtration, the witness complex,  is constructed such that it possess the same topological properties as the original one.  A well-known method for selecting landmark points, is the \textit{maxmin} algorithm. The methodology of the maxmin algorithm and the construction of  the witness complex can be found in detail in \cite{Silva2004}. We make use of the implementation available in \cite{javaPlex}.
 %~~~~~~~~~~~~~~~~~~~~~~~~~~~~~~~~~~~~~~~~~

%~~~~~~~~~~~~~~~~~~~~~~~~~~~~~~~~~~~~~~~~~~~

%~~~~~~~~~~~~~~~~~~~~~~~~~~~~~~~~~~~~~~~~~
%  \begin{figure}[tb]
%      \centering
%   \includegraphics[width=0.24\linewidth]{figs/empty.png}
%    \includegraphics[width=0.24\linewidth]{figs/empty.png}
%     \includegraphics[width=0.24\linewidth]{figs/empty.png}
%      \caption{ \small {Env + complex + Persistent Diagrams for a couple of runs}}
%      \label{sim2}
%   \end{figure}
 %~~~~~~~~~~~~~~~~~~~~~~~~~~~~~~~~~~~~~~~~~

\section{Encounter Metric }\label{enc-met}
%On the other hand, for a network of moving agents, instead of analyzing data as a collection of points $(p_i(t), t)$, with $p_i(t)$ referring to the position vector of the $i'$th agent at time $t$,  we can make use of  only a subsample of which, the data obtained from encountering of the agents during their movement. This enables us to build a filtered complex that is not dependent on the positioning but the encountering data, which is called encounter complex. 
%Consider a mobile sensor network in which to each agent  a unique ID has been assigned.
% Denote the set of IDs assigned to  all agents by $\mathcal{I}$. 
  Let $\mathcal{I}$ 
%  $= \{1,2, ..., n\}$ 
  denote the set of all of the IDs assigned to sensing nodes.    
%  We assume that  each sensing node is able to detect the IDs of its neighboring nodes within the radius of $R_d$, the detection radius. When a node detects another node within its detection radius, an \textit{encounter event} is recorded and sent to the base station. 
For moving nodes in the network,  an
\textit{encounter event} $\E_i$ occurs at time $t_i$ if   $\exists\, I_i^1, I_i^2 \in \mathcal{I} $ such that $\|p(I_i^1, t_i) - p(I_i^2, t_i) \| \leq r_d $, where $p: (\mathcal{I},R) \mapsto \mathbb{R}^2$ is a coordinate function such that $ p(i, t) $ is the position vector of the node $i$ at time $t$. The encounter $\E_i$ is represented as a tuple
\begin{equation}
  \E_i= [t_i, I_i^1, I_i^2  ],
\end{equation}
and its corresponding  ID set is defined as  $I_i = \{ I_i^1, I_i^2 \}$.
To build a distance metric on the set of encounter events,  we construct  an undirected weighted graph $\G$ with vertices corresponding to the events $\E_i$, denoted as \textit{encounter graph}. For each two vertices $\E_i$ and $\E_j$, they are considered connected if $I_i \cap I_j \neq \emptyset$, and disconnected otherwise. The condition $I_i \cap I_j \neq \emptyset$ implies that there exist a node $k\in \I$ which has encountered two other agents at times $t_i$ and $t_j$ at positions $p_i$ and $p_j$, respectively. Due to the sensing limitations, these coordinates are not available at the base station. However, the Euclidean distance between $p_i$ and $p_j$ is bounded by $u_{ij} = v_m . |t_i- t_j|$, where $ v_m = max(v_c, v_p) $. Therefore, one can assign a weight proportional to $u_{ij}$ as a rough estimation of $\| p_i - p_j \|$ to the undirected  edge  connecting $\E_i$ and $\E_j$ as
$
w_{ij} =
\begin{cases}
   |  t_i -t_j|,& \text{if } \I_i \cap \I_j \neq \emptyset\\
 \infty,              & \text{otherwise}
\end{cases}
$.  Now we build a metric on $\G$,  denoted by $\D_{\G}$, 
%whose elements are defined as the length of the shortest path between the corresponding nodes in $\G$, i.e. $
as $[ \D_\G]_{i,j} =    d_\G(\E_i,\E_j) $
where $d_\G(\E_i,\E_j)$ represents the length of the shortest path between nodes $  \E_i,\E_j$ in $ \G $.
%~~~~~~~~~~~~~~~~~~~~~~~~~~~~~~~~~~~~~~~~~
  \begin{figure}[tb]
      \centering
   \includegraphics[width=0.99\linewidth]{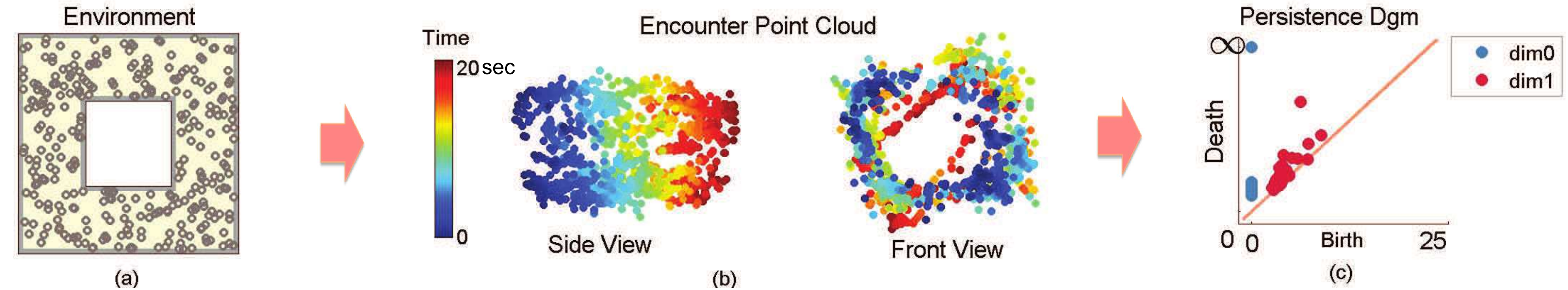}
      \caption{ \small {(a) An environment with one hole, (b) the encounter point cloud for data gathered over 20 seconds and (c) the persistence diagram}}
      \label{encmds}
   \end{figure}
%~~~~~~~~~~~~~~~~~~~~~~~~~~~~~~~~~~~~~~~~~
  \begin{figure}[b]
      \centering
   \includegraphics[width=1\linewidth]{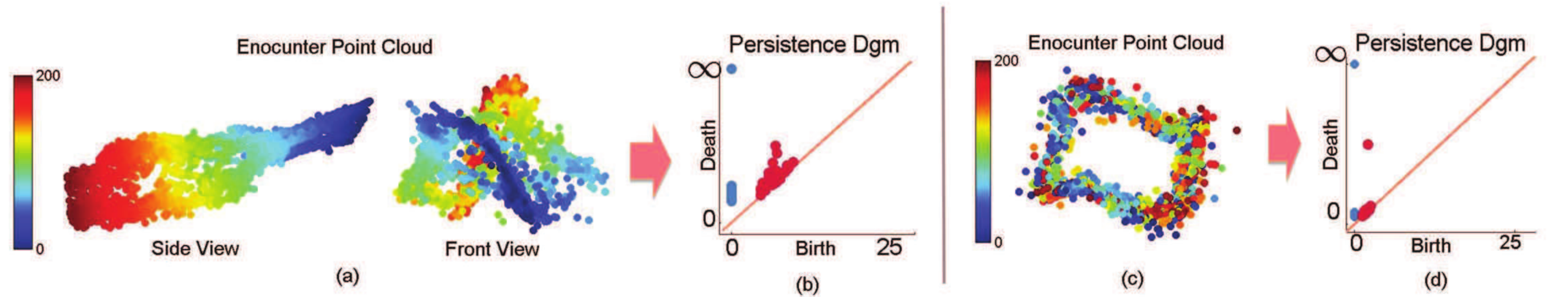}
         \caption{ \small {Comparison of the point clouds obtained from encounter data collected over 200 seconds using (a) encounter metric in the mobile network, (c) using contracted metric on the hybrid network, and their corresponding persistence intervals in (b) and (d), respectively.
%         point cloud for a hybrid network, and (d-f) the corresponding persistence intervals over 200 sec. 
}}
      \label{encImpr}
   \end{figure}
%~~~~~~~~~~~~~~~~~~~~~~~~~~~~~~~~~~~~~~~~~
\subsection{ Encounter Tracking}
%\noindent\textbf{Encounter Tracking:}
Due to the fact that the agents can have probabilistic stops in our model, taking into account every single encounter detection will result in redundant data. Therefore, we  consider  the encounters that take place only at the beginning of a period when two agents are in proximity with each other for the whole interval. Specifically, 
the encounter event $\E_i$ occurs at time $t_i$ if  (i)  $\exists I_i^1, I_i^2 \in \mathcal{I} $ such that  $\|p(I_i^1, t_i) - p(I_i^2, t_i) \| \leq r_d $, and (ii)  $\exists \, t'\in [0,  t_i] $ such that $ \|p(I_i^1, t) - p(I_i^2, t) \| > r_d $ for $\forall t \in [t', t_i )$.

 Figure \ref{encmds} illustrates an example of a square environment with one hole inside on which 200 agents perform exploration. The encounter point cloud, colored over time of simulation, (figure \ref{encmds}(b)) resembles a noisy cylindrical tube, and one can easily distinguish the hole in the corresponding dgm$_1(X) $ (figure \ref{encmds}(c)). 
 However, as we continue data collection for a longer time interval, the cycles in its two ends tend to get deformed and the point cloud loses its cylindrical shape (figure \ref{encImpr}(a)). In this case, the hole is not distinguishable from noise in the corresponding  dgm$_1(X)$ (figure \ref{encImpr}(b)).
 We can justify this due to the accumulation of the errors in the encounter metric because of the uncertainty in pairwise distances over a long period of time, which makes the estimated point cloud gradually lose the topological structure  existing in the real environment.
 Note that as it could be observed from figure \ref{enc}(b), these encounters produce  sampled points in the space of $ \D \times [0, t_f]$, and collecting more observation over time does not help the estimation of $ \D $ but it makes it worst. 
One solution to this problem would be to consider point clouds obtained from information over a shorter time frame that is still long enough to capture the correct structure. However, in practice, this would impose the requirement of tuning another parameter. In the following  we will show that by adding an extra piece of information one can build more precise \textit{contracted} point clouds that carry correct topological information without worrying  about time frames. 

\subsection{{Contracted Encounter Metric}}
To increase the precision of estimated pairwise distances in our metric, we consider the fact that the agents stop probabilistically for some time intervals,  and keep track of stopped nodes in the network by asking the agents to transmit their change of moving status (RW, WF, or S). If two nodes meet the third one during one of its stop intervals, we infer the corresponding encounters have occurred at the same location. 
Hence we use this observation and improve the metric as follows:
Let $ \mathcal{T}_k  = \{T_k^1, T_k^2, \dots\}$ be the set of stop intervals
%  $T^{}_{s} = [t_{{}}^{s}, t^e_{{}}]$ be one of the  stop intervals 
  for agent $k$, meaning that agent $k$ has been in S mode for $\forall \, t \in  T_k^{l}, \, \forall l $.  For two events $\E_i$ and $\E_j$, 
if  $I_i \cap I_j =  k  $ and $ t_i, t_j \in  T^{l}_{k}$ for some $l$, then we set $w_{ij} =0.$ We refer to this metric as a \textit{contracted encounter metric} as it contracts the space of the corresponding encounter point cloud  from $ \D \times [0, t_f] $ to the lower dimension space of $ \D $.
We observed that contracting the point cloud improves the performance of the estimation in terms of more persistence features, but it still is not a complete embedding into a lower dimensional space as a result of the probabilistic nature of stop intervals of the agents. 
% of   feature extraction from  the real environment as it contracts 
%. Figure \ref{encImpr}(b) and (e) present the results for a contracted metric for the same environment and time interval as the \ref{enc-met}  example, respectively.
%%in figure  and the data gathered over the same time interval. 
%The 1D prominant feature is easily distinguishable from noise in the persistence diagram.[\tc{red}{discussion}].

\subsection{A Hybrid Network}
We can also make further improvements over the estimated metric by incorporating a few percentage of  static nodes in the  network, which we refer to  it as a \textit{hybrid network}. The exploration task initiates with all nodes starting in a RW status, and after a while, when the nodes are dispersed enough in the environment, a few percentage of them are commanded  to stop as static nodes. 
Now, let $\mathcal{S} \in \mathcal{I}$ be the set of indices for these static  nodes. Then we modify the corresponding weights for two encounter events $\E_i$  and $\E_j$ in case that $ \I_i \cap \I_j \neq \emptyset$ as  
% $w_{ij} = 0$ if  $I_i \cap I_j \in \mathcal{S}$. 
%$
%w_{ij} =
%\begin{cases}
%   \|  t_i -t_j\|,& \text{if } \I_i \cap \I_j \neq \emptyset \text{ and } I_i \cap I_j \notin \mathcal{S}\\
%      0,& \text{if } \I_i \cap \I_j \neq \emptyset \text{ and } I_i \cap I_j \in \mathcal{S}\\
% \infty,              & \text{otherwise}
%\end{cases}
%$. 
$
w_{ij} =
\begin{cases}
   \|  t_i -t_j\|,& \text{if }  I_i \cap I_j \notin \mathcal{S}\\
      0, & \text{if }  I_i \cap I_j \in \mathcal{S}\\
\end{cases}
$. 
This configuration can be used in case that the natural stop intervals of agents are not long enough to contract the point cloud space into a proper lower dimensional space. 
The resulting contracted point cloud and persistence diagrams for a hybrid network with only $\% 5 $ of static nodes are shown in figure \ref{enc-met}(c) and (d), which nicely infer the existence of the environment by a much better separation between the persistent feature and the noise

%\section{(corner detection ?)}

\section{Subsampling }\label{subs}

In \cite{Dirafzoon2013}, we used the well-known maxmin landmark selection  algorithm \cite{Silva2004} for selecting a subset of point cloud on which the complexes are built.  
%This algorithm is an   inductive selection process, in which 
% the first landmark $l_1$ is chosen  randomly from the point cloud  $V$ equipped with a metric $d$.  Iteratively, if $L=\{l_1, ..., l_{k-1}\}$ is the set of previously selected  landmarks, find the next landmark as a point $l_{k} \in V $ s.t. maximizes $d(l_k,L)$.
   Unfortunately,  maxmin filtration is very sensitive to outliers as they are distant from the other points in the set and very prone to be selected as the maximizers of the distance to the set of previously selected samples. 
%        $d(l_k,L)$. 
Dealing with the outliers in real-world data analysis is of  high importance. One of the causes of the appearance of outliers in the estimated data sets is due to the estimation uncertainties. In our problem, they occur as a result of  inaccuracies in the estimation of encounter metric, which can be observed in figures \ref{enc-met} and \ref{encImpr}.
%...  helps extraction of features in presence of noise - robust ] ... \\
%observation: existence of outliers in data due to ... \\
 We propose two  approaches to overcome this issue and  improve the robustness of our mapping algorithm by enhanced filtration of the point cloud. 
 
\subsection{{KNN filtering}}
 In \cite{Carl07} nearest neighbor density estimation  is used for  pre-filtration  of the point cloud. In our first approach, we  employ  a k-nearest neighbors (KNN) filtration followed by a maxmin subsampling algorithm to select a subsample of point cloud which preserves the topological information of the actual space and is robust to outliers. Specifically,  for a point $ x_i \in {X} $, let $ d_k(x_i) $ denote the distance between $x_i$
 and it's k-nearest neighbor, and $ \overline{d}_k(x_i) $ be the average distance  to its k-nearest neighbors (inversely proportional to the local density of the point cloud around $ x_i $). We consider the density of all average  distances over ${X}$  as $\rho_k ({X})$ and select the threshold $\tau_{q,k}$ to be $q$-quantile of $\rho_k ({X})$. 
 Then we select a subset $V$ of points $ v_i $ such that $d_k(v_i) <\tau_{q,k}, \, \forall v_i \in V $, which will remove outliers of ${X}$  to a large extent. Finally, we apply the maxmin algorithm on the pre-filtered point cloud $V$ to obtain a smaller subset $V_s$ for persistence analysis. 
%\tc{cyan}{
%filtration has been used for 
%-  Thresholding based on distances of KNN neighbors
% - Random subset selection 
% - Subset selection based on a probability density (density of k-nearest neighbors vs. )
% }
%~~~~~~~~~~~~~~~~~~~~~~~~~~~~~~~~~~~~~~~~~
  \begin{figure}[tb]
      \centering
          \includegraphics[width=0.85\linewidth]{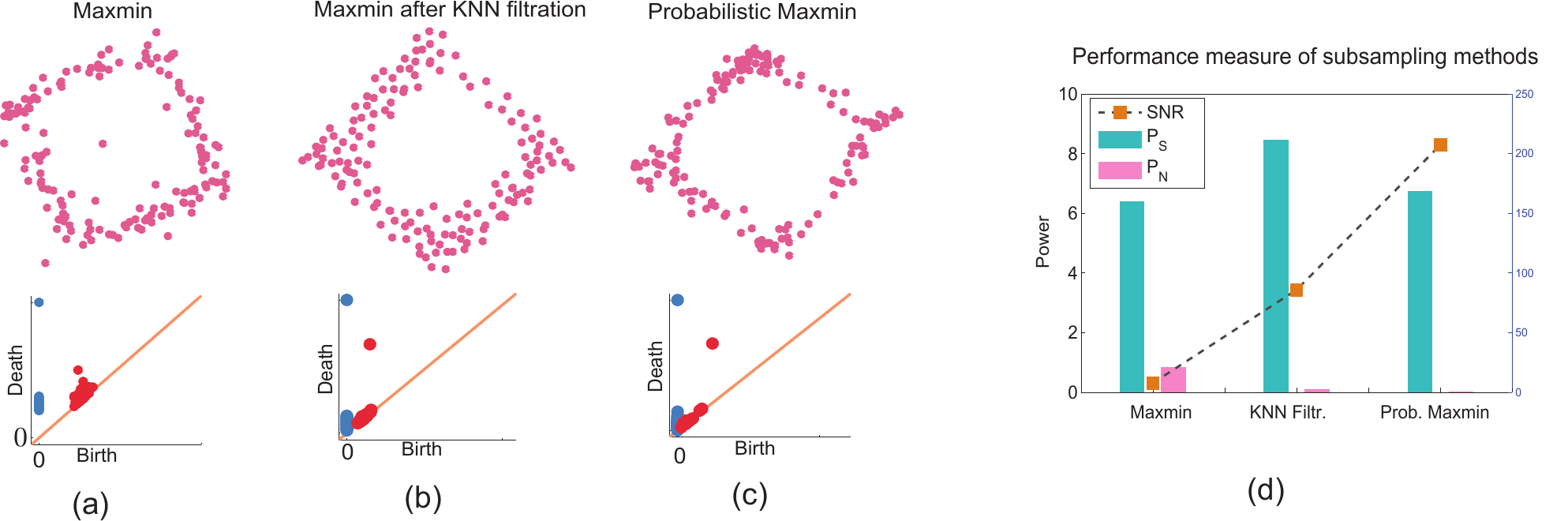}
      \caption{ \small{Comparison of (a) standard maxmin subsampling,  (b) maxmin over pre-filtered data, and (c) probablistic maxmin algorithm, and (d) their SNR performance for 100 independent runs  }} 
      \label{land2}
   \end{figure}
 %~~~~~~~~~~~~~~~~~~~~~~~~~~~~~~~~~~~~~~~~~ 

\subsection{{Probabilistic maxmin}}
In the second approach, we propose a single stage sequential probabilistic subsampling method, where the local densities of the point cloud are taken into account directly in sequential sample selection procedure rather than in a separate pre-filtration process. We select  the first sample $x_1 \in{X}$  randomly; Then at each iteration, if $L$ is the set of the previously selected landmarks, find a point $l_{k}\in {X}$ such that it maximizes $\alpha(d(l_{k},L), \rho_k(l_k))$, where $ \alpha(u,v) $ is a scalar continuous function with the properties:  $\alpha(0,0) = 0$, and  $\alpha(u,.)$ is an increasing  function of $ u $ for a fixed $ v $,  $\alpha(.,v)$ is an increasing  function of $ v $ for a fixed $ u $. An appropriate example of such a function for our application is $\alpha(u,v)  = u + \omega v$ where $ \omega $ is a constant scalar. 

Figure \ref{land2}(a-c) shows the subsampled point clouds of 150 points for a  contracted data set in along with their persistence diagrams with parameters $ q=0.9 $ and $ k=10 $. It confirms that the one dimensional feature is more persistent using KNN filtration or our probabilistic maxmin subsampling than the standard maxmin approach. As another example, a subsampled point cloud from encounter information over a 2-hole square environment projected on 2D space is shown in figure \ref{env}. The points are colored according to the real positions of encounters in the environment, confirming that it nicely represents the structure of the  environment. 
In order to compare the performance of landmark selection algorithms in presence of noise, we use a signal to noise ratio (SNR) measure as follows: For a point  $(b^i_n, d^i_n) $  in  dgm, let 
$l^i_n = [b_n^i, d_n^i]$ be the corresponding persistence interval. 
Then $| l^i_n| $
%$l^i_n = |b_n^i -d_n^i|$ 
represents the vertical  distance of the point to the diagonal of dgm , and can be considered as a rough measure of significance or noisiness of features. In other words, features with small values of $ |l^i_n| $ can be considered as noise and the ones with larger values as signals, with some level of confidence \cite{Bal13}. 
%  of the feature . If the point is close enough to the diagonal ()  it can be interpreted either as noise or as signal.
Let $ \mathcal{L}_n  = \{|l^1_n| , ..., |l^m_n|\}$ be the set of all interval lengths  in dimension $ n $ sorted in non-decreasing order.  Define the signal set as $ \mathcal{S}_n = \{|l^i_n| \in \mathcal{L}_n\,| i< \beta_n+1 \} $ and the noise set as the set difference $ \mathcal{N}_n  = \mathcal{L}_n \backslash \mathcal{S}_n$. Then their corresponding average powers can be defined as 
$  P_{\mathcal{S}_n}  = \frac{ \sum_{\mathcal{S}_n} 
{|l_n^i|}^2}{ |\mS_n|}  $ 
%\frac{l^{i^2}_n}{|\mS_n|}$ 
and 
%$ P_{\N_n}  =  \sum_{\N_n} \frac{l^{j^2}_n}{|\N_n|}$, 
$ P_{\N_n}  = \frac{ \sum_{\N_n}  {|l^j_n|}^2}{ |\N_n|}$, 
and the signal to noise ratio as $ SNR = \frac{ P_{\mS_n} }{P_{\N_n} }  $. A number of $ N $ independent subsampling runs can then be performed over the point cloud, and the average SNR can be used as the performance measure of the subsampling method. 
The average SNR measure for the scenario in figure \ref{encmds} after 100 runs is shown in figure \ref{land2}(d)exposing a great improvement of SNR for these approaches over the standard maxmin algorithm. 

%~~~~~~~~~~~~~~~~~~~~~~~~~~~~~~~~~~~~~~~~~
\section{Robust Classification  of Persistence Intervals}\label{class}
%~~~~~~~~~~~~~~~~~~~~~~~~~~~~~~~~~~~~~~~~

Betti numbers summarize topological features of a space $M$ as numbers.
%but are not stable[].
 However, for a sampled point  cloud $X$ out of $M$, one usually constructs a filtration of simplicial complexes on $ X $ and analyzes varying homology  over a scale space. For a dense enough sampled data, one can find a range of scales for which the homology of simplicial complexes is equal to the one for $M$ \cite{Bal13}. Nonetheless, 
%Each scale, $\epsilon$ is usually represented as a (distance) function over the point cloud. We usually start with a zero $\epsilon$ which results to as many connected components as the number of samples and finish with a scale close to the max distance between the points, which glues everything together as a single component with no holes. 
%Hence, the homology of the sets in the filtration would be equal to the one for the actual space only for a certain range of scales. 
%That is one motivation behind using persistence intervals instead of betti numbers for the analysis of  varying features over a filtration. 
 persistence intervals do not provide a  quantitative representation of the topology of the real space but only birth and death of the features over the scale space. 
%  represented as a persistence diagram or barcode (see figure \ref{clust-ex}). 
%In this paper, we are interested to extract quantitative description of features in the topological space based on the persistence intervals for  homologies over the filtration. 
Furthermore, we are interested in feature extraction algorithms that are robust to scaling and outliers. 
%As an example, for three sampled point clouds in figure \ref{clust} (a),  the length of the prominent 1-dimensional feature (shown in figure \ref{clust} (b)) varies by changing the radius or presence of the outliers, but the spaces from which  these point clouds are sampled have the same topological features. 
% Another issue in the analysis using persistence diagrams is that they are not scale-invarient. 
%  Consider, for example the two circles shown in figure \ref{clust-ex} with different radii and their corresponding persistence diagrams,  in which the length of the prominent hole varies by increasing the radius. However, these two circles topologicaly are equivalent. Furthermore, the presence of outliers in the sampled data,  alters the persistence of dominant features (see figure \ref{clust-ex}). 
%  
In the following part, we propose a density based classification algorithm for  persistence intervals with the purpose  of a scale invarient and robust feature extraction method for sampled data sets.  
%~~~~~~~~~~~~~~~~~~~~~~~~~~~~~~~~~~~~~~~~
\begin{figure}[tbp]
\begin{center}
\includegraphics[width=1\linewidth]{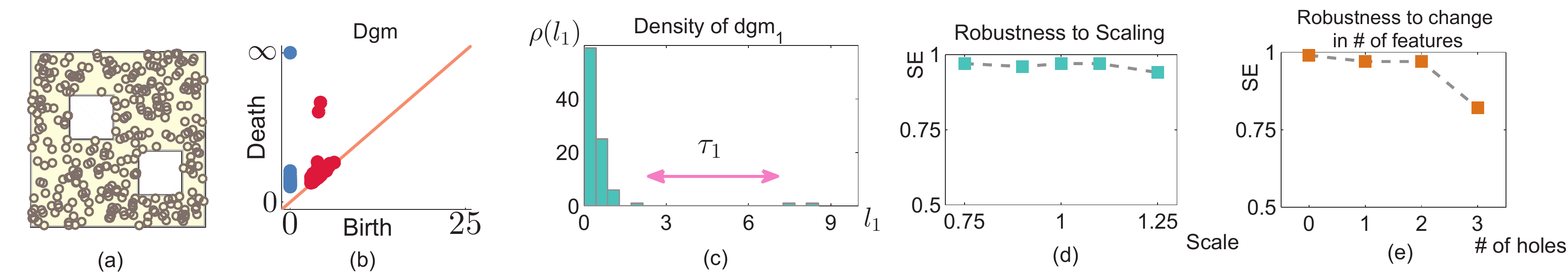} 
\end{center}
\caption{ \small{
% (a) Sampled point cloud from unit circle (green), samples from the same circle with added outliers  (blue), and sampled data from a larger circle (red), (b) comparison of dgm$ _1$ for three point clouds, (c) 
% 
 Robust Classification of persistence intervals: (a) A 2-hole environment, (b) the extracted persistence diagram, (c) density of persistence intervals with the appropriate threshold,  (d) robustness of classification algorithm to scaling,  and (e)  to change in  the number of features }}  \label{clust}
\end{figure}
%~~~~~~~~~~~~~~~~~~~~~~~~~~~~~~~~~~~~~~~~~
%\begin{table}[!b]\label{tab1}
%\caption{{Performance analysis} }
%\label{per}
%\begin{center}
%\begin{tabular}{cc}
%\begin{tabular}{c|c|c|c|c|c}
%Scale & 0.75  & 0.9   & 1 & 1.1 & 1.25  \\
% \hline 
%SE       & 		&&&& \\
%\end{tabular} & 
%\begin{tabular}{c|c|c|c|c}
%1D cycles & 0   & 1   & 2 & 3   \\
% \hline 
%SE       & 		&&&\\
%\end{tabular} 
%\end{tabular}
%\end{center}
%\end{table}
%%~~~~~~~~~~~~~~~~~~~~~~~~~~~~~~~~~~~~~~~~~

Consider again the persistence interval lengths $| l^i_n |$. 
We are interested in finding a threshold for dimension $ n $,  $ \tau_n $ such that all of the features with the property that $| l^i_n |>   \tau_n$  can be considered as  signals (see figure \ref{clust}(d-e)). To make the approach scale invariant, we consider the density of interval lengths scaled by a $ q_n $ quantile, $ \rho_{q_n} (l_n)$, and define the $n$-th Betti function as 
%$
%\hat{\beta}_n (q_n, \Delta_n, \tau_n)= \sum_i{\mathds{1}    \left(     \frac{l^{i}_n}{l_{q_n} + \Delta_n }> \tau_n   \right)},$
$
\hat{\beta}_n (q_n, \Delta_n, \tau_n)= \sum_i{\mathds{1}_{\mathbb{R}^+}    \left(     \frac{l^{i}_n}{l_{q_n} + \Delta_n } - \tau_n   \right)},$
 where $l_{q_n}$ is the $ q_n $-quantile of the density, 
$\Delta_n$ has been added for dealing with singularities, and  $\mathds{1}_{\mathbb{R}^+} (.)  $ is the indicator function of ${\mathbb{R}^+}$ .
%[\tc{red}{there exist parameters for which $ \hat{\beta}_n (q_n, \Delta_n, \tau_n) =  \beta_n $}]. 
%In order to  find the optimal  values for parameters, we employ bootstrapping  method. 
Let $\mathbb{M} = \{M_1, \ldots, M_m\}$  denote a set of random spaces  that  have the same topological characteristics as $M$, and $\mathbb{X} = \{X_1, \ldots, {X_m}\}$ represent the corresponding set of sampled point clouds. Define the $(n,k)$-th Betti function as 
%\[
%\hat{\beta}^k_n (\Theta_n ) = \sum_i{\mathds{1}    \left(     \frac{l^{i,k}_n}{l_{q_n}^{k} + \Delta_n }> \tau_n   \right)}. 
%\] 
\[
\hat{\beta}^k_n (\Theta_n ) = \sum_i{\mathds{1}_{\mathbb{R}^+}     \left(     \frac{l^{i,k}_n}{l_{q_n}^{k} + \Delta_n } - \tau_n   \right)}, 
\] 
and its corresponding error function as 
$
e_n^{k} (\Theta_n ) = | \hat{\beta}^{k}_n (q_n, \Delta_n, \tau_n)-  \beta_n|
$,
where 
 $\Theta_n = (q_n, \delta_n, \tau_n)$ defines the parameter space. 
Now our classification problem reduces to perform  an optimization algorithm to find the minimum of the cost functional $\mathcal{F}(\Theta_n) = \left\| [e_n^{1}(\Theta_n), ... , e_n^{m}(\Theta_n)]^T \right\|_1/m$, i.e. the optimal parameter set is 
%$
%{e}_n(\Theta_n )= [e_n^{1}, ... , e_n^{m}]^T
%$
$ 
\Theta^*_n  = \argmin_{\Theta_n}  \mathcal{F}(\Theta_n)  
$.
%$
%\Theta^*_n  = \argmin_{\Theta_n} || {e}_n(\Theta_n ) ||_1   
%$
%\[
%\hat{\beta}^j_n = \sum_i{\mathdsm{1}    \left(     \frac{l^{i}_n}{l_{q^j_n}^{i} + \Delta^j_n }> \tau^j_n   \right)}. 
%\]
%Training:
%$\Theta^j_n = (q^j_n, \delta^j_n, \tau^j_n)$
%m independent runs \\
%estimated $\beta_n$ using $\Theta^j_n$ for the k-th run denoted by $\hat{\beta}^{j,k}_n$
%Define the error
%\begin{equation}
%e_n^{j,k} = | \hat{\beta}^{j,k}_n -  \beta_n|
%\end{equation}
%
%\begin{equation}
%\overline{e}^j_n= [e_n^{j,1}, ... , e_n^{j,m}]^T
%\end{equation}
%
%\begin{equation}
% f(\Theta_n^j) = e^j_n=| \overline{e}^j_n|_1
%\end{equation}
%\begin{equation}
%\Theta^*_n  = \argmin_{\Theta^j_n} f(\Theta_n^j)   
%\end{equation}
%[\tc{red}{TODO: Discrete optimization]}
To make the cost functional  smooth for optimization purposes, one can replace the indicator function $ \mathds{1}(.) $
with a sigmoid function $ \sigma_{\alpha ,l^i_n}(.) $ where 
$\sigma_{\alpha, x_0} (x) = \frac{1}{1+e^{-\alpha(x-x_0)}}$.
%\begin{equation}
%\hat{\beta}^j_n = \sum_i{\sigma_{ ... ,l^i_n}    \left(     \frac{l^{i}_n}{l_{q^j_n}^{i} + \Delta^j_n }> \tau^j_n   \right)}
%\end{equation}

We performed classification for dimension $ 1 $ intervals on a training data set consisting of 100 variations of the space shown in figure \ref{clust}(a) with random placements of the two square holes in the space. Persistence diagram and the corresponding density of interval lengths for one of the environments is shown in figure \ref{clust}(b) and (c), respectively. 
The optimization process resulted in a minimum of  $ \mathcal{F} (\Theta^*_1)  = 0.03$ with  $\Theta^*_1 = (0.5 , 0.7,  3.7)$.  This result is not surprising as the $0.5$-quantile of the density, which is the median of dataset, is known to be robust to outliers. 
To investigate the performance of our classifier, we evaluated $\mathcal{F} (\Theta^*_1)$ for a variety of  test sets $ \mathbb{X}^i $, each containing  100 point clouds with random feature placements, with each test set differing in either scale of the environment and features or the number of features (holes). 
To investigate the performance, we used  the sensitivity measure for classification, $ SE  = TP/ (TP+FN)$, where $ TP$ and $ FN $ denote the number of true positives and false negatives, respectively. The sensitivity performance for 
these scenarios is summarized in figure \ref{clust} (d) and (e) for scaled environments, and environments with different numbers of holes but the same scale, respectively. Note that we have evaluated this measure for the whole mapping algorithm and not just for the classification part, which can justify the degradation in the performance for the 3-hole case, as we observed that in $15\%$ of the cases, the third hole could not be retrieved from the point cloud data. 

\section{Conclusion}\label{conc}
We introduced encounter metrics for the construction of point clouds that represent topological features of unknown environments based on minimal encounter information of mobile nodes in a sensor networks whose mobility is inspired by insects. We enhanced the accuracy of the estimation with incorporating different pieces of information in the construction of the metric. Moreover, we employed density based subsampling approaches to cope with the outliers emerging due to uncertainties in the point cloud, and proposed a classification  for robust feature extraction out of persistence diagrams. 
Future work includes validation of the proposed algorithms on swarm robotic and biobotic systems as well as the quantification of uncertainties in topological estimation. We are also working on how one can extract  features  other than connected components and holes from the encounter point cloud can be exploited for a more precise mapping technique.

%ACKNOWLEDGMENTS are optional
%\section*{Acknowledgments}
%Acknowledgement goes here.

\bibliographystyle{ieeetr}
%\bibliography{SoCG14.bbl}

%\begin{thebibliography}{}

%\end{thebibliography}
%
%\begin{thebibliography}{1}
%
%\bibitem{IEEEhowto:kopka}
%H.~Kopka and P.~W. Daly, \emph{A Guide to \LaTeX}, 3rd~ed.\hskip 1em plus
%  0.5em minus 0.4em\relax Harlow, England: Addison-Wesley, 1999.
%
%\end{thebibliography}

% that's all folks
\end{document}